\documentclass[]{aastex7}
\usepackage{multirow}
\usepackage{amssymb} 
\usepackage{booktabs}
\usepackage{color}

\usepackage{comment}
\graphicspath{{./}{Figures/}}

\begin{document}

\title{ComptonUNet: A Deep Learning Model for GRB Localization with Compton Cameras under Noisy and Low-Statistic Conditions}

\author[orcid=0000-0003-3070-5880,sname='Sato']{Shogo Sato}
\affiliation{Faculty of Science and Engineering, Waseda University}
\affiliation{NTT Corporation}
\email[show]{shg.sato@ntt.com}

\author[orcid=0000-0001-6916-9654,sname='Tanaka']{Kazuo Tanaka}
\affiliation{Faculty of Science and Engineering, Waseda University}
\email{tanaka@kaduo.jp}

\author{Shojun Ogasawara}
\affiliation{Faculty of Science and Engineering, Waseda University}
\email{shojun02-oubutsu@suou.waseda.jp}

\author{Kazuki Yamamoto}
\affiliation{Faculty of Science and Engineering, Waseda University}
\email{kazuki-onepiece@akane.waseda.jp}

\author[orcid=0000-0001-7697-9575,sname='Murasaki']{Kazuhiko Murasaki}
\affiliation{NTT Corporation}
\email{kazuhiko.murasaki@ntt.com}

\author[orcid=0000-0002-5379-3150,sname='Tanida']{Ryuichi Tanida}
\affiliation{NTT Corporation}
\email{ryuichi.tanida@ntt.com}

\author[orcid=0000-0003-2819-6415,sname='Kataoka']{Jun Kataoka}
\affiliation{Faculty of Science and Engineering, Waseda University}
\email{kataoka.jun@waseda.jp}


\begin{abstract}
Gamma-ray bursts (GRBs) are among the most energetic transient phenomena in the universe and serve as powerful probes for high-energy astrophysical processes. In particular, faint GRBs originating from a distant universe may provide unique insights into the early stages of star formation. However, detecting and localizing such weak sources remains challenging owing to low photon statistics and substantial background noise. Although recent machine learning models address individual aspects of these challenges, they often struggle to balance the trade-off between statistical robustness and noise suppression. Consequently, we propose ComptonUNet, a hybrid deep learning framework that jointly processes raw data and reconstructs images for robust GRB localization. ComptonUNet was designed to operate effectively under conditions of limited photon statistics and strong background contamination by combining the statistical efficiency of direct reconstruction models with the denoising capabilities of image-based architectures. We perform realistic simulations of GRB-like events embedded in background environments representative of low-Earth orbit missions to evaluate the performance of ComptonUNet. Our results demonstrate that ComptonUNet significantly outperforms existing approaches, achieving improved localization accuracy across a wide range of low-statistic and high-background scenarios. 
\end{abstract}


\section{Introduction}~
Gamma-ray bursts (GRBs) are among the most energetic transient phenomena in the universe and provide critical insights into the deaths of massive stars, neutron star mergers, and the early evolution of the cosmos \citep{piran2004physics, woosley2006supernova}. Since the 1990s, missions such as the Burst and Transient Source Experiment (BATSE) aboard the Compton Gamma Ray Observatory (CGRO) satellite have enabled the detection of thousands of GRBs, revealing their isotropic distribution and establishing their extragalactic origins \citep{paciesas1999fourth}. More recently, the Swift satellite, launched in 2004, provided rapid GRB localization and multi-wavelength follow-up with its Burst Alert Telescope (BAT), X-Ray Telescope (XRT), and Ultraviolet/Optical Telescope (UVOT), leading to major breakthroughs in understanding GRB progenitors and their afterglows \citep{gehrels2004swift}. However, the detection and accurate localization of weak and short GRBs are still challenging primarily because of limited photon statistics under background contamination. These faint events, which often originate from the most distant universe, provide rare opportunities for probing the early stages of star formation. Consequently, the development of advanced techniques for accurately estimating gamma-ray directions under these challenging conditions is highly desired.

Although traditional GRB detectors such as BATSE employed large scintillator arrays with wide fields of view (FoV) and high sensitivity, recent advances in miniaturized satellite platforms, particularly small satellites, have enabled the low-cost and high-frequency deployment of high-energy detectors in space \citep{serjeant2020future, werner2018camelot}. One such mission is the INnovative Space Probe for Imaging R-process Emission (INSPIRE) with a Compton camera, developed by Waseda University and the Institute of Science Tokyo, and scheduled for launch in 2027 \citep{kataoka2024inspire, tanaka2024mev}. The wide field of view of the INSPIRE CC-Box, which covers approximately $1/4$ of the entire sky, makes it highly suitable for observing not only steady sources but also unpredictable transient events such as GRBs, black hole binaries, and active galactic nuclei (AGN) flares. In particular, the accurate GRB localization is expected to play a crucial role in multi-messenger astronomy by coordinating with gravitational wave observations from LIGO and Virgo~\citep{abbott2017gw170817, abbott2020prospects}. However, compared to large-scale predecessors like BATSE, the microsatellite-based INSPIRE mission is constrained by a limited detector size, making it difficult to accumulate sufficient photon statistics within such short timeframes. To address these limitations, INSPIRE aims to detect weak GRBs through the incorporation of deep learning techniques tailored to the platform’s limited detection area and the short duration of these events.

Recently, deep learning has been increasingly applied to Compton camera data analysis to enable accurate image reconstruction under challenging conditions \citep{daniel2022application, yao2022rapid}. A common approach involves convolutional neural networks (CNNs) such as Unet \citep{ronneberger2015u} to estimate the gamma-ray directions from the reconstructed Compton image \citep{sato2020high}. This approach is inherently robust to background noise, because event selection during the reconstruction process removes most of the background events, and the model is trained to reduce background noise. However, this event selection may discard events that implicitly contain information regarding gamma-ray directions, resulting in degraded performance under limited photon statistics. More recently, ComptonNet \citep{sato2024comptonnet} was proposed, which directly processes raw data using a neural network to estimate gamma-ray directions. This approach is particularly effective with limited statistics, making it suitable for transient detection. However, because ComptonNet treats all measured events, including background and noise, as valid inputs, its performance may degrade significantly in high-noise environments dominated by the cosmic X-ray background (CXB) \citep{boldt1987cosmic} or atmospheric albedo \citep{stephens2015albedo}.

Consequently, we propose ComptonUNet, a hybrid deep learning framework that jointly processes raw data in ComptonNet manner and reconstructed images in Unet manner. By leveraging the complementary strengths of both models, ComptonUNet enables accurate GRB localization even under limited photon statistics and significant background contamination. We simulated GRB-like events under realistic space radiation backgrounds and compared its performance with those of existing models to evaluate ComptonUNet's performance.

The remainder of this paper is organized as follows. Section 2 describes the configuration of the Compton camera, the architectures of the proposed models with its baselines, and the simulation setup used for dataset generation. Section 3 presents the comparative results of the reconstruction performance. Section 4 discusses the noise robustness of the model, the contribution of each input component via an ablation study, and the astrophysical implications of the results. Finally, Section 5 concludes the paper. Detailed implementation parameters of the neural networks are provided in Appendix A.

\section{Materials and methods} 
\subsection{Compton camera configuration and reconstruction technique}
\begin{figure*}[ht!]
\epsscale{0.5}
\plotone{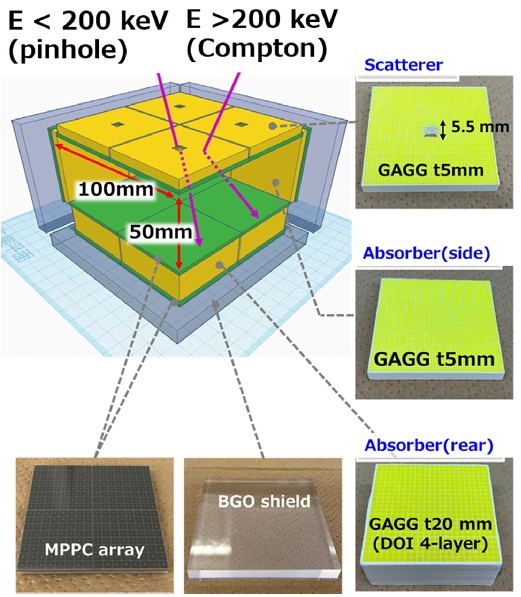}
\caption{\label{f0} Detailed configuration of the CC-BOX (Compton Camera Box) onboard the INSPIRE satellite~\citep{kataoka2024inspire}. The figure illustrates the front Ce:GAGG pixel array, side and bottom BGO scintillator active shields, pinhole structure, and multilayer detector arrangement. This design enables high-sensitivity and high-angular-resolution observations over a wide energy range (30~keV–3~MeV).}
\end{figure*}

The primary gamma-ray detector aboard the INSPIRE satellite is the Compton-camera box (CC-Box), which is a compact, lightweight instrument designed for MeV gamma-ray observations \citep{kataoka2024inspire,tanaka2024mev}, as depicted in~Figure~\ref{f0}. The CC-Box integrates both pinhole and Compton imaging capabilities, covering the energy range from 30~keV to 3~MeV. The effective FoV of the CC-Box is approximately 3.1~sr, enabling the detection of transient phenomena such as GRBs. Its design was optimized for high sensitivity and angular resolution within the constraints of a 50~kg-class satellite platform. 

The CC-Box comprises a 2~$\times$~2 array of sensor modules, each comprising a pixelated cerium-doped gadolinium aluminum gallium garnet (Ce:GAGG) scintillator optically coupled to a 16~$\times$~16-channel multi-pixel photon counter (MPPC) array. The front layer is a 5~mm-thick Ce:GAGG array with 1~$\times$~1~mm$^2$ pixels, featuring a central 5.5~$\times$~5.5 mm$^2$ pinhole for low-energy imaging. The rear section includes four layers of 20~mm-thick Ce:GAGG arrays with 2~$\times$~2~mm$^2$ pixels, forming a depth-of-interaction (DOI) structure to improve spatial resolution. Additional Ce:GAGG elements are positioned on the sides to enhance off-axis sensitivity. The CC-Box is equipped with bismuth germanate (BGO) scintillator shields surrounding its sides and bottom to suppress background radiation and identify incomplete energy deposition events. Each BGO shield is read out by a monolithic MPPC, providing active shielding and aiding in background discrimination.

For image reconstruction in Compton mode, the CC-Box utilizes Compton kinematics to determine the gamma-ray directions. The scattering angle $\theta$ can be calculated using the Compton formula by measuring the energy deposits and interaction positions in the two layers,
\begin{equation}
\cos \theta = 1 - \frac{m_e c^2}{E_2} + \frac{m_e c^2}{E_1 + E_2},
\end{equation}
where $E_1$ and $E_2$ are the energy deposits in the scatter and absorber layers, respectively, and $m_e c^2$ is the electron rest mass energy. Each event defines a ``Compton cone" representing the possible gamma-ray directions, and we reconstruct images by superimposing Compton cones. In pinhole mode, gamma-ray directions are calculated using a pinhole principle that relates the detection position to the incoming direction. Conventional reconstruction methods, such as back projection (BP) and maximum likelihood expectation maximization (MLEM), integrate multiple Compton cones or pinhole events to visualize the spatial distribution of gamma-ray incidences. However, these methods require a large number of events for accurate reconstructions, making them less effective under low-statistic conditions, which is typical for short-duration or weak-flux GRBs. Consequently, we investigated deep learning based reconstruction methods capable of operating under low-statistics and background conditions, thereby enhancing the performance of compact Compton cameras likesuch as the CC-Box for weak-GRB observations.

\subsection{Machine Learning Models}
We adopted deep-learning models for GRB localization to overcome the limitations of traditional image reconstruction under low-statistics and high-background conditions. In this section, we introduce two established architectures, Unet \citep{ronneberger2015u} and ComptonNet \citep{sato2024comptonnet} as baseline models. We then propose a novel hybrid framework, ComptonUNet, which integrates the strengths of both models.

\begin{figure*}[ht!]
\plotone{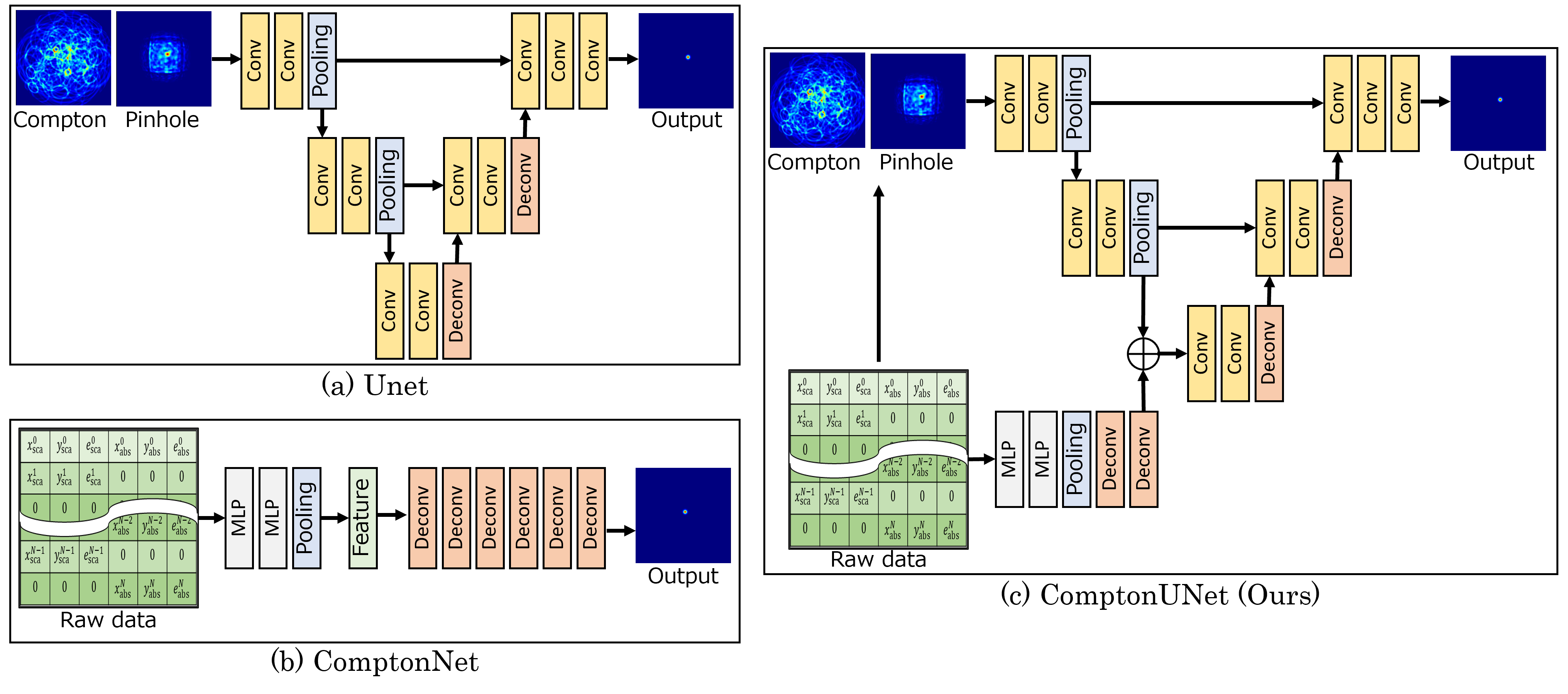}
\epsscale{1.2}
\caption{\label{f1} Overview of (a) Unet, (b) ComptonNet, and (c) ComptonUNet architectures. Unet processes reconstructed images, while ComptonNet directly estimates gamma-ray directions from raw data. ComptonUNet combines both approaches by transforming raw data into feature maps using a ComptonNet-inspired encoder, followed by Unet-style decoder to estimate the gamma-ray directions. This design improves robustness under low-statistics and high-background conditions, making it suitable for transient gamma-ray observations.}
\end{figure*}

Unet, originally developed for biomedical image segmentation, has been widely adopted for image reconstruction. In the context of GRB localization with Compton cameras, Unet-based models process reconstructed images using Compton and pinhole modes to estimate gamma-ray directions. The reconstructed images were calculated through event selection based on the energy deposits and respective BP algorithms. The architecture comprised an encoder-decoder structure with skip connections, which allows it to learn hierarchical features and perform effective denoising, as shown in Figure~\ref{f1} (a). This approach provides robust performance in noisy environments by implicitly learning to denoise the input data. However, Unet-based models rely on reconstructed images from sufficient events and tend to degrade their performance when the number of input events is severely limited.

In contrast, ComptonNet is a specialized neural network designed to directly estimate gamma-ray directions from raw data without requiring intermediate image reconstruction. Here, the "raw data" consists of normalized feature vectors with 16 channels containing energy depositions and interaction coordinates ($E, x, y, z$) from each detector segment (front, rear, side, and BGO shields), hence ComptonNet can effectively utilize the full statistical information available from the detector. The ComptonNet comprises shared-weight multilayer perceptrons (MLPs), max pooling, and deconvolution layers, as shown in Figure~\ref{f1} (b). This approach provides robust performance under low-statistic conditions by directly mapping raw data to the gamma-ray directions. However, ComptonNet is highly sensitive to background contamination, because it treats all measured events as valid inputs. In high-background environments, such as those dominated by CXB and albedo, the model can struggle to distinguish between signal and noise, leading to degraded performance.

To address these complementary limitations, we propose ComptonUNet, a hybrid deep learning framework that jointly processes raw data and reconstructed images for robust GRB localization. As shown in Figure~\ref{f1} (c), this model first transforms raw data into feature maps with a ComptonNet-inspired encoder and then applies the Unet-style convolutional model to estimate the gamma-ray directions. This design allows the model to tolerate both limited statistics and significant noise contamination, making it well-suited for space-based transient gamma-ray observations such as GRBs. In this study, following ComptonNet, we selected the mean squared error (MSE) for the loss function during training. We trained and evaluated all models using simulated data that incorporated both signal and realistic background sources, and compared their performance across a range of GRB durations. Detailed specifications of these input formats and the reconstruction logic are provided in Appendix A.

\subsection{Dataset preparation}
We generated a synthetic dataset using Geant4~\citep{agostinelli2003geant4} simulations in which the full geometry and material properties of the CC-Box were faithfully reproduced to train and evaluate the proposed machine learning models under realistic observational conditions. The detector configuration, including the Ce:GAGG scintillators, MPPC, and shielding components, was modeled based on the engineering specifications described in \citep{kataoka2024inspire}.

GRB events were simulated as point sources toward the CC-Box with a fixed photon flux of 1.0~photons~cm$^{-2}$~s$^{-1}$, which corresponds to the typical value reported in GRB population studies \citep{paciesas1999fourth}. Simulations were performed with burst durations of 1, 3, 10, 30, and 100~s to evaluate model performance across a range of photon statistics. For each duration, we generated 1000 independent simulation runs, each with a GRB source placed at a random position within the 30$^\circ$ half-angle FoV of the CC-Box. To train and evaluate the models, we split the dataset into training (640 runs), validation (160 runs), and test (200 runs) sets. The energy spectra of the GRB sources were assumed to follow a power-law distribution with a photon index $\Gamma$ randomly sampled from the range of $[-2.5, -1.3]$, which is consistent with the observed GRB spectral slopes in the 50~keV – 10~MeV range. Each simulation included background contributions from both the CXB and the albedo. Following the empirical fit \citep{gruber1999spectrum}, we adopted a simple power-law approximation of the CXB and albedo spectrum with a photon index of $\Gamma = -2.88$ and $\Gamma = -1.35$, respectively.

All photons were tracked through the full detector geometry to produce event-level data including energy deposits, reaction positions, and timing information. Energy and position resolutions were considered in this simulation, following the CC-BOX. No pre-selection or event filtering was applied prior to model training, to preserve the statistical and noise characteristics of realistic GRB observations.

\subsection{Quantitative evaluation metrics}
We employed a set of quantitative metrics commonly used in image analysis as well as specialized criteria tailored to GRB localization to assess the performance of each reconstruction model.

The overall reconstruction accuracy was evaluated using two standard metrics: MSE and structural similarity index measure (SSIM)~\citep{wang2004image}. The MSE quantifies the pixel-wise differences between the reconstructed image and the corresponding ground truth, providing a measure of the global deviation. SSIM, in contrast, evaluates the perceived structural similarity between the predicted images and ground truth by considering luminance, contrast, and texture information. In addition to general performance metrics, we utilized a task-specific criterion for GRB localization: the angular offset between the predicted and ground-truth gamma-ray directions. For each image, the predicted direction was estimated by computing the intensity-weighted centroid of the reconstructed image. The angular separation between the centroid and ground-truth directions was then calculated. This ``peak offset" directly quantifies localization accuracy and serves as a key indicator of a model’s suitability for GRB localization. Following the ComptonNet paper, we trained each model ten times independently for each experimental condition (i.e., for each GRB duration). In these runs, the dataset split was fixed, but the weight initialization and data shuffling were varied using different random seeds. For quantitative assessment, we selected the top five runs that achieved the lowest MSE on the validation dataset and computed the mean and variance of their evaluation metrics using the test dataset. This approach ensures that the reported results are robust against random initialization and training fluctuations, thereby providing a reliable estimate of the model performance under each condition.

\begin{figure*}[ht!]
\plotone{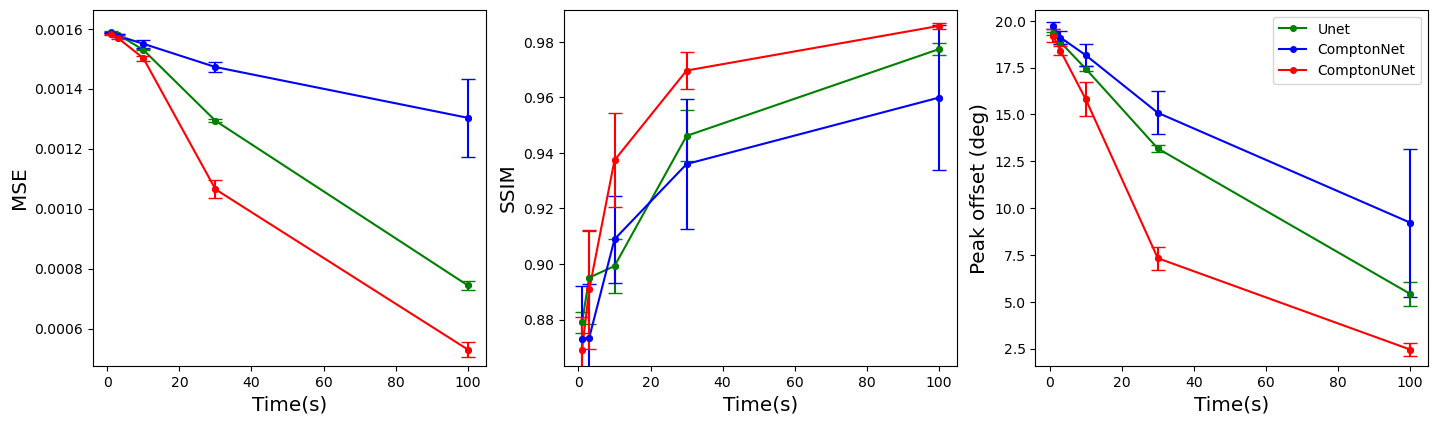}
\caption{\label{f2} Summary of quantitative performance for the three models: (a) MSE, (b) SSIM, and (c) peak offset. ComptonUNet outperforms both Unet and ComptonNet across all metrics, by combining the strengths of both architectures.}
\end{figure*}

\begin{figure*}[ht!]
\epsscale{0.8}
\plotone{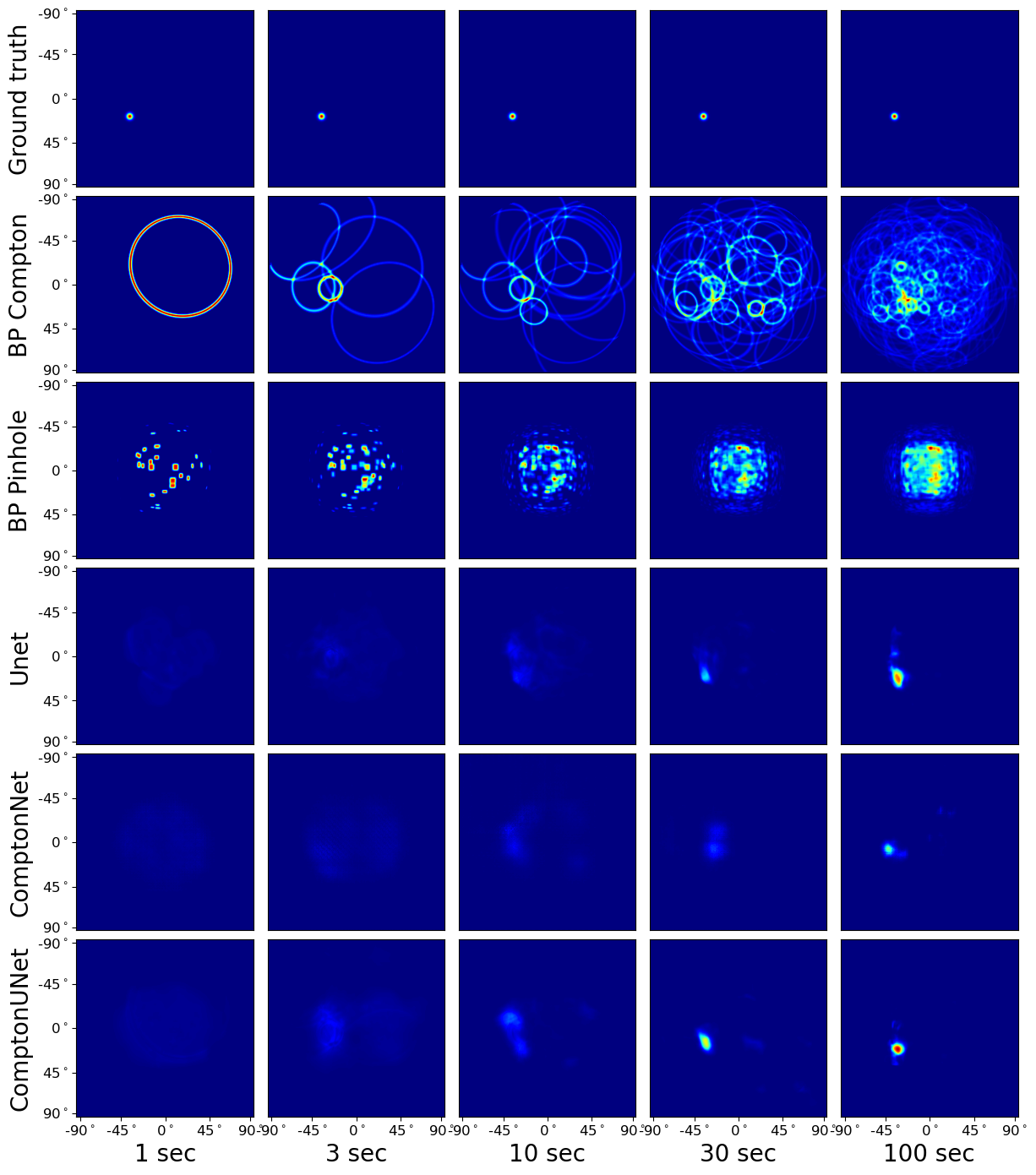}
\caption{\label{f3} Visual comparison of reconstructed images from the three models. From top to bottom: Ground truth, BP image of Compton mode (BP Compton), BP image of pinhole mode (BP Pinhole), Unet output, ComptonNet output, and ComptonUNet output. Additionally, from left to right, the images represent different GRB durations (1, 3, 10, 30, and 100~s). The ComptonUNet model demonstrates superior performance in accurately reconstructing the source morphology and peak location compared to the other models.}
\end{figure*}

\section{Results}
\subsection{Quantitative and visual performance comparison}
We first compared the performances of the three models, Unet, ComptonNet, and ComptonUNet (ours), using the quantitative metrics described in Section~2.4. Figure~\ref{f2} summarizes the results in terms of the MSE, SSIM, and peak offset across all test cases. For MSE, all models exhibited improved estimation accuracy with increasing GRB duration. In the 1–10~s range, accurate localization remains difficult because of severely limited photon statistics, resulting in low signal amplitudes and a high MSE across all models. For durations beyond 30~s, the photon statistics are still insufficient for Unet, and the background effects are still critical for ComptonNet. In contrast, ComptonUNet consistently achieves lower MSE values, demonstrating its ability to balance statistical efficiency with noise robustness. The SSIM values exhibit 0.88--0.90 among the models for short durations (1--10~s), as all models produce low pixel values close to zero due to the severely limited photon statistics. For durations exceeding 30~s, ComptonUNet demonstrates superior performance, mirroring the trend observed in the MSE results. Similarly, the peak offset follows a trend comparable to that of the MSE, with ComptonUNet consistently providing more accurate localization across most conditions. Consequently, even with the limited detector size of a 50kg-class satellite, ComptonUNet achieves a localization accuracy of $7.5^\circ$ at 30~s and $2.5^\circ$ at 100~s assuming a fixed flux of 1.0~photons~cm$^{-2}$~s$^{-1}$. Note that, the geometrical area of INSPIPRE is only $\sim$1/20 of one of eight NaI(Tl) scintillators constituting the BATSE detector.

Because ComptonNet directly estimates gamma-ray directions from raw data without using reconstructed images, and includes background events, its training process is unstable and the inference performance varies significantly depending on the initialization parameters. In contrast, ComptonUNet incorporates reconstructed images as additional inputs, which provide analytically calculated gamma-ray directions. This enables ComptonUNet to achieve greater stability and reduced variance in its predictions compared with ComptonNet.

For comparison, we also evaluated the localization accuracy of the traditional BP method. The peak offsets for BP were 20.7$^\circ$, 19.7$^\circ$, 16.3$^\circ$, 14.8$^\circ$, and 13.9$^\circ$ for durations of 1, 3, 10, 30, and 100~s, respectively. While the BP method shows some improvement with duration, the accuracy is significantly inferior to that of ComptonUNet, which achieves 2.2$^\circ$ at 100~s. Note that we do not compare MSE or SSIM for BP, as these pixel-wise metrics are heavily biased by image sparsity in low-statistic regimes and do not correctly reflect localization performance in this context.

Figure~\ref{f3} presents representative output images from each model under identical simulation conditions. The ground truth, BP images, and model predictions are shown for the GRB durations of 1, 3, 10, 30, and 100~s. In Compton-mode BP, only a few cones are visible for 1–3~s, which makes localization difficult. At 100~s, the GRB directions are roughly distinguishable. In pinhole-mode BP, the GRB source is located outside the FoV; consequently, it does not appear in the reconstructed images. The Unet performance improves with increasing duration, particularly from 30 to 100~s, but still exhibits residual noise owing to the limited photon counts. ComptonNet performs worse than Unet under background environments; the localized position remains deviated from the ground truth even at 30~s, and the source position becomes distinguishable only at 100~s. In contrast, ComptonUNet recovers the GRB peak direction with less noise, outperforming both baseline models. Whereas Unet tends to oversmooth the signal and ComptonNet introduces noise artifacts, ComptonUNet provides a robust compromise, confirming the trends observed in the quantitative evaluations.

\begin{figure*}[ht!]
\plotone{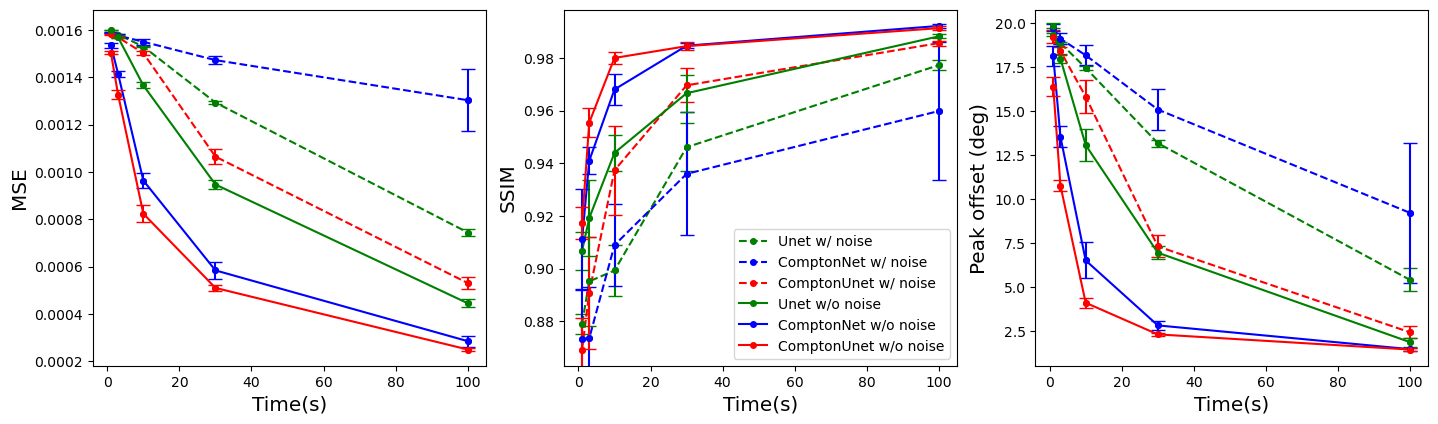}
\caption{\label{f4} Noise robustness evaluation of the three models under eliminated background conditions. The dashed lines and simple lines represent the performance of each model under background conditions (as shown in Figure~\ref{f2}) and eliminated background conditions, respectively. ComptonUNet maintains its performance even in the presence of background noise, while ComptonNet shows significant degradation.}
\end{figure*}

\section{Discussion}
\subsection{Noise Robustness}
In this section, we evaluate the noise robustness of ComptonUNet by comparing its performance with those of Unet and ComptonNet under different background conditions. Specifically, we assess the model performance with and without background contributions from the CXB and atmospheric albedo, allowing us to isolate the impact of noise on the measured events.

Figure~\ref{f4} shows the results in terms of MSE, SSIM, and peak offset for all models under both background-present and background-removed scenarios. As expected, the performance of all models improves when background noise is eliminated. Unet exhibits relatively stable performance under both conditions, as much of the background is removed during the event selection processes. In contrast, ComptonNet exhibits a substantial degradation in performance under background-present conditions, primarily because it processes all measured events, including noise, without explicit background filtering. ComptonUNet, however, demonstrates a considerably smaller performance gap between background and no-background conditions. This indicates that its hybrid architecture effectively mitigates the influence of noise, balancing statistical sensitivity with denoising capability.

\begin{figure*}[ht!]
\epsscale{0.8}
\plotone{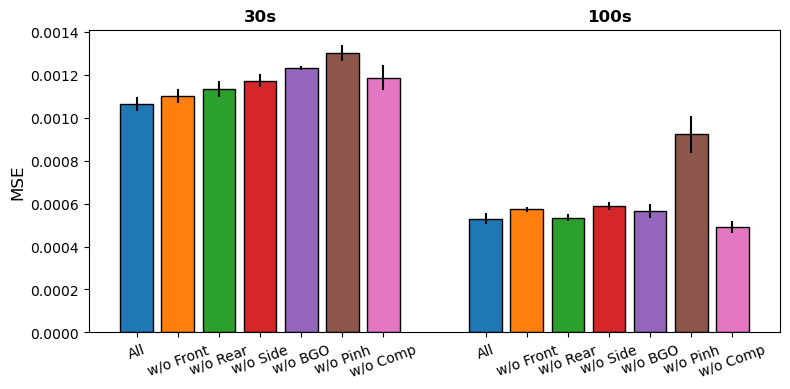}
\caption{\label{f5} Ablation study of ComptonUNet. The ablation study demonstrates the importance of each component in the ComptonUNet inputs for MSE evaluation. "All" represents the full input. Additionally, w/o Front, Rear, Side, BGO, Pinh, and Comp ignored the front, rear, side, BGO, pinhole image, and Compton image components, respectively. The results indicate that the pinhole image plays a crucial role in improving the performance of ComptonUNet, since it provides stable hints of gamma-ray direction through analytical reconstruction based on the pinhole principal.}
\end{figure*}

\subsection{Ablation study}
We conducted an ablation study in which each component was removed in turn to assess the contribution of the individual input components to the performance of ComptonUNet. ComptonUNet receives raw data from multiple detector segments (front, rear, side, and BGO), and reconstructs the images (pinhole mode and Compton mode). We evaluated the impact on the reconstruction performance by systematically omitting each component individually.

Figure~\ref{f5} presents the results of this ablation analysis, showing the mean MSE and standard deviation (error bars) derived from the top five independent training runs for each configuration. The figure indicates that the full ComptonUNet model achieves the best performance at both 30~s and 100~s. In the 30~s analysis, removing the pinhole image component results in a significant performance degradation, highlighting its critical role in providing stable hints of the gamma-ray direction through analytical reconstruction based on the pinhole principle. Additionally, the BGO component plays an important role by providing gamma-ray direction constraints from count ratios and removing escape events via veto signals. At 100~s, the pinhole image remains significant for source localization. \footnote{Although the "w/o Comp" configuration appears to have a slightly lower mean MSE at 100~s compared to the full model, this difference falls within the standard deviation, suggesting that it is attributable to statistical fluctuations rather than a physical advantage.}

\subsection{Comparison with BATSE Observations}
BATSE aboard the CGRO achieved remarkable success in detecting various GRBs with different fluxes and durations. BATSE comprised eight large-area NaI(Tl) scintillation detectors, each with an effective area of approximately 2025~cm$^2$, mounted on the corners of the satellite to provide nearly all-sky coverage. The system employed simple count-rate threshold triggers and triangulation techniques to localize GRBs, with typical positional uncertainties of several degrees depending on the photon statistics and incident angle. Owing to its large detection area and wide FoV, BATSE could detect thousands of GRBs, including several short and faint events.

We compared the localization accuracy of BATSE with that of the CC-Box on INSPIRE combined with ComptonUNet to contextualize the performance of our proposed system. We selected BATSE as a benchmark because it shares a similar observation energy band and wide-field survey objective, serving as a suitable reference for degree-level localization unlike coded-mask missions (e.g., Swift). 

Specifically, we focused on GRBs with peak fluxes in the range of 0.8–1.2~photons~cm$^{-2}$~s$^{-1}$, which is consistent with our simulation setup. We compared the positional uncertainty reported in The 4B BATSE GRB catalog~\citep{paciesas1999fourth} with the peak offset distribution derived from our simulations, using the inter-quartile range (IQR; 25th–75th percentile) as a metric. For GRBs with $T_{90} \sim 30$~s, BATSE exhibited a positional uncertainty of 3.04--5.57$^\circ$, whereas INSPIRE with ComptonUNet yielded a peak offset range of 3.08--9.03$^\circ$. At $T_{90} \sim 100$~s, BATSE achieved 1.80--2.26$^\circ$, whereas ComptonUNet produced a comparable IQR of 1.02--2.32$^\circ$. It should be noted that these ComptonUNet performances differ from the mean performance shown in Figure 3. While Figure 3 presents the average of the top five training runs to demonstrate model stability, the values reported here are derived from the single best-performing model among them, assuming a realistic deployment scenario where the optimal model is selected for operation.

Notably, this comparison is not strictly equivalent. The BATSE results were derived from real observational data, whereas the INSPIRE+ComptonUNet results are based on detailed Geant4 simulations. Additionally, the BATSE values represent estimated uncertainties rather than measured offsets from true positions, as ground truth is unavailable for most BATSE events. Moreover, BATSE was specifically designed for transient detection, whereas the CC-Box on INSPIRE was optimized for steady gamma-ray sources and did not employ a geometry specialized for GRB localization. Furthermore, the CC-Box has a significantly smaller effective detection area ($\sim$100~cm$^2$) than BATSE's $\sim$2000~cm$^2$, which inherently limits its photon collection capabilities. Nonetheless, the comparison provides a useful benchmark, suggesting that advanced deep learning–based reconstruction enables competitive localization performance, even with compact instruments operating under constrained photon statistics.

\subsection{Limitations}
Despite these advantages, several limitations of this study must be acknowledged. 

First, it relied entirely on simulated data. Although the simulations incorporate realistic background sources, such as CXB and albedo, the generalization of ComptonUNet to real-world conditions, including detector response nonuniformities, calibration uncertainties, and hardware-related artifacts, remains to be evaluated. The incorporation of experimental data from laboratory setups or orbital missions is essential for validating the practical performance of a model.

Second, the GRB-like sources in this study were restricted to within 30$^\circ$ of the CC-Box FoV center, which is consistent with the FoV limits of 30$^\circ$ for the pinhole mode and 60$^\circ$ for the Compton mode. However, the actual GRB sky distribution is isotropic, and the model performance may vary significantly in off-axis regions. Future work should explore the robustness of the model across a wider range of gamma-ray directions and angles, particularly using the BGO count rates such as BATSE.

Third, the size of the training dataset (1,000 samples per duration) was selected to verify the proposed method while maintaining reasonable computational costs. We confirmed the sufficiency of this size for stable training by conducting 10 independent runs, which yielded consistent results with small variance as shown in Figure 3. However, it is possible that further improvements in localization accuracy could be achieved by significantly increasing both the dataset size and the model capacity (i.e., the number of parameters). Exploring such large-scale models, potentially at the cost of higher computational requirements, remains an important direction for future research.

\subsection{Scientific Implications}
In the context of multi-messenger astronomy, a localization accuracy of approximately $3^\circ$--$5^\circ$ is the practical threshold for coordinating effective follow-up observations. Our results with ComptonUNet demonstrate that effective localization achieving this accuracy is feasible for long GRBs (e.g., 100~s) even under typical fluxes (1.0~photons~cm$^{-2}$~s$^{-1}$) using a microsatellite-scale detector. Regarding short GRBs, while detection remains challenging under typical flux conditions due to limited photon statistics, the proposed method could potentially be applicable to exceptionally bright events where the integrated photon counts are comparable to those of long GRBs. Precise position determination further enables the application of Angular Resolution Measure (ARM) cuts~\citep{kataoka2025revisiting}. By filtering events based on the derived source direction, the signal-to-noise (S/N) ratio can be significantly improved, thereby facilitating accurate spectral extraction even for faint sources.

\subsection{Strategy for Application to Real Observations}
In this study, we evaluated models trained on discrete durations and a fixed flux to demonstrate the feasibility of the proposed method. For application to real observational data, where GRBs exhibit continuous variations in duration and flux, we propose a ``Model Bank'' strategy. In this approach, a library of network weights is pre-trained on a grid of conditions covering various durations and average fluxes. In the analysis pipeline, the system first detects a GRB trigger and estimates its approximate duration ($T_{90}$) and average flux from the light curve. Subsequently, the model weights trained on the condition closest to the observed parameters are selected for inference. Furthermore, since the inputs to our model are time-integrated images and count vectors, the system is robust to the temporal variability of the flux as long as the integrated photon statistics align with the selected model's training domain.

\section{Conclusion}
We presented ComptonUNet, a hybrid deep learning architecture for estimating gamma-ray direction using Compton camera data under low-statistic and high-background conditions. ComptonUNet demonstrated robust performance across varying GRB durations and background environments by combining the event-level inference capability of ComptonNet with the denoising strengths of U-Net. Through detailed Geant4-based simulations of the CC-Box instrument aboard the upcoming INSPIRE mission, we demonstrated that ComptonUNet consistently outperformed baseline models in terms of reconstruction accuracy, noise robustness, and localization precision. These results highlight the potential of hybrid deep learning approaches for enhancing transient gamma-ray detection using compact space-based instruments. Future work will focus on validating the model performance with experimental and in-orbit data to further advance the applicability of deep learning techniques in high-energy astrophysics.

\begin{acknowledgments}
This research was supported by the Japan Science and Technology Agency (JST) ERATO Grant No. JPMJER2102.
\end{acknowledgments}

\begin{contribution}
Shogo Sato conducted the core research activities, including the development of the ComptonUNet model, implementation of simulations, data analysis, and manuscript preparation.
Kazuo Tanaka, Kazuhiko Murasaki, and Ryuichi Tanida contributed to the conceptual development of the study through technical discussions, and provided interpretation of results, particularly in the Discussion section.
Shojun Ogasawara and Kazuki Yamamoto assisted in the construction and preprocessing of simulation datasets used in the model training and evaluation phases.
Jun Kataoka served as the principal investigator of the project, securing research funding, guiding the overall scientific direction, and providing continuous feedback on the methodology and manuscript content.
All authors reviewed and approved the final manuscript.
\end{contribution}

\appendix
\section{Implementation Details}
This paper presents a novel hybrid deep learning architecture, ComptonUNet, for gamma-ray direction estimation using Compton camera data. This appendix provides additional implementation details for ComptonUnet and the baseline models to facilitate reproducibility and further understanding of the model.

\smallskip\noindent
\textbf{Unet}
The Unet baseline is implemented as an encoder-decoder network with skip connections. The encoder consists of two stages (depth=2) of convolutional blocks. Each block includes two $3 \times 3$ convolutional layers, each followed by Batch Normalization and a LeakyReLU activation. The input consists of two channels ($256 \times 256$ pixels). For this study, the Unet model was configured as a lightweight version with 18,817 trainable parameters. During training, a batch size of 16 was used.

\smallskip\noindent
\textbf{ComptonNet}
ComptonNet processes raw event data directly. The encoder comprises shared-weight MLPs, transforming the 16-channel input features into a 128-dimensional latent space. The decoder consists of five transposed convolutional layers. This model contains 3,855,120 trainable parameters and was trained with a batch size of 4.

\smallskip\noindent
\textbf{ComptonUNet} ComptonUNet integrates the encoders of both architectures. The raw data path uses two NonLinear layers (64 and 128 units), while the image path processes reconstructed images through three convolutional stages (32, 64, and 128 filters). These paths are concatenated at the bottleneck (256 channels). The total number of trainable parameters is 5,550,593. Similar to ComptonNet, it was trained with a batch size of 4.

\smallskip\noindent
\textbf{Shared Training Details and Computational Cost} All models were implemented using PyTorch and trained on an NVIDIA GeForce RTX 2080 Ti GPU. The training process employed the Adam optimizer with the MSE loss function used as the primary objective for all architectures. To ensure optimal generalization and prevent overfitting, an early stopping strategy was implemented; specifically, training was terminated if the validation loss failed to improve for 30 consecutive epochs after an initial warm-up period of 50 epochs. During this process, the model weights corresponding to the minimum validation loss were preserved for final evaluation. 
The computational efficiency was assessed by monitoring peak GPU memory consumption and processing time during training with the 100~s duration dataset. Under these conditions, the Unet model required 4.37~GB of GPU memory, while ComptonNet and the proposed ComptonUNet utilized 1.28~GB and 2.50~GB, respectively. Regarding computational time, Unet was the fastest, with an average training time of 8.54~s per epoch (21.6~min total for 152 epochs). ComptonNet required 30.4~s per epoch (41.5~min total for 82 epochs). As expected, the proposed ComptonUNet was the most demanding, requiring 31.9~s per epoch (50.0~min total for 94 epochs). The relatively small difference in per-epoch time between ComptonNet and ComptonUNet implies that the processing of raw event data dominates the computational load. For reference, the inference time for processing 200 test samples was 1.13~s (5.65~ms/sample) for Unet, 7.49~s (37.5~ms/sample) for ComptonNet, and 7.72~s (38.6~ms/sample) for ComptonUNet.

\smallskip\noindent
\textbf{Input Data (Raw data)} The input to the ComptonNet-based encoder consists of raw event-level information derived from the CC-Box. For each detected event, we construct a 16-channel feature vector that encapsulates the energy deposition recorded in the Front, Rear, and Side GAGG scintillators, as well as the BGO active shield. Furthermore, the vector includes the three-dimensional interaction coordinates $(x, y, z)$ and its deposit energy $E$ for each event. These features are normalized to a $[0, 1]$ range before being processed by the shared-weight MLP blocks. In cases where multiple interactions occur within the same detector type, the position corresponding to the highest energy deposition is selected as the interaction location, and the total energy deposition from all interactions is summed to represent the event's energy.

\smallskip\noindent
\textbf{Input Data (Reconstructed images)} The image-based path of ComptonUNet and the Unet baseline receive two types of $256 \times 256$ pixel images representing different detection modes. The first type is the pinhole image, generated by utilizing the front detector layer as a pinhole camera; hit positions on the rear detector are back-projected through the physical aperture of the CC-Box to the sky plane, effectively capturing low-energy photons. The second type is the Compton image, used for multi-hit events identified as Compton scattering. A simple Compton BP algorithm calculates the scattering angle $\theta$ from energy depositions using the Compton formula, projecting possible source directions as cones onto the sky map. The final image is formed by accumulating these cones, which provides spatial constraints for higher-energy photons. To enhance the quality of the reconstructed images, we use front, rear, and side GAGG detector data for image reconstruction, while employing the BGO active shielding to veto incomplete energy deposition events, thereby reducing background noise.

\bibliography{sample7}{}

\begin{thebibliography}{}
\expandafter\ifx\csname natexlab\endcsname\relax\def\natexlab#1{#1}\fi
\providecommand{\url}[1]{\href{#1}{#1}}
\providecommand{\dodoi}[1]{doi:~\href{http://doi.org/#1}{\nolinkurl{#1}}}
\providecommand{\doeprint}[1]{\href{http://ascl.net/#1}{\nolinkurl{http://ascl.net/#1}}}
\providecommand{\doarXiv}[1]{\href{https://arxiv.org/abs/#1}{\nolinkurl{https://arxiv.org/abs/#1}}}

\bibitem[{B.~P. Abbott {et~al.}(2017)Abbott, Abbott, Abbott, Acernese, Ackley, Adams, Adams, Addesso, Adhikari, Adya, {et~al.}}]{abbott2017gw170817}
Abbott, B.~P., Abbott, R., Abbott, T.~D., {et~al.} 2017, \bibinfo{title}{GW170817: observation of gravitational waves from a binary neutron star inspiral,} Physical review letters, 119, 161101

\bibitem[{B.~P. Abbott {et~al.}(2020)Abbott, Abbott, Abbott, Abraham, Acernese, Ackley, Adams, Adya, Affeldt, Agathos, {et~al.}}]{abbott2020prospects}
Abbott, B.~P., Abbott, R., Abbott, T., {et~al.} 2020, \bibinfo{title}{Prospects for observing and localizing gravitational-wave transients with Advanced LIGO, Advanced Virgo and KAGRA,} Living reviews in relativity, 23, 3

\bibitem[{S. Agostinelli {et~al.}(2003)Agostinelli, Allison, Amako, Apostolakis, Araujo, Arce, Asai, Axen, Banerjee, Barrand, {et~al.}}]{agostinelli2003geant4}
Agostinelli, S., Allison, J., Amako, K.~a., {et~al.} 2003, \bibinfo{title}{{GEANT4—a simulation toolkit},} Nuclear instruments and methods in physics research section A: Accelerators, Spectrometers, Detectors and Associated Equipment, 506, 250

\bibitem[{E. Boldt(1987)Boldt}]{boldt1987cosmic}
Boldt, E. 1987, \bibinfo{title}{The cosmic X-ray background,} Physics Reports, 146, 215

\bibitem[{G. Daniel {et~al.}(2022)Daniel, Gutierrez, \& Limousin}]{daniel2022application}
Daniel, G., Gutierrez, Y., \& Limousin, O. 2022, \bibinfo{title}{Application of a deep learning algorithm to Compton imaging of radioactive point sources with a single planar CdTe pixelated detector,} Nuclear Engineering and Technology, 54, 1747

\bibitem[{N. Gehrels {et~al.}(2004)Gehrels, Chincarini, Giommi, Mason, Nousek, Wells, White, Barthelmy, Burrows, Cominsky, {et~al.}}]{gehrels2004swift}
Gehrels, N., Chincarini, G., Giommi, P., {et~al.} 2004, \bibinfo{title}{The Swift gamma-ray burst mission,} The Astrophysical Journal, 611, 1005

\bibitem[{D. Gruber {et~al.}(1999)Gruber, Matteson, Peterson, \& Jung}]{gruber1999spectrum}
Gruber, D., Matteson, J., Peterson, L., \& Jung, G. 1999, \bibinfo{title}{The spectrum of diffuse cosmic hard X-rays measured with HEAO 1,} The Astrophysical Journal, 520, 124

\bibitem[{J. Kataoka {et~al.}(2024)Kataoka, Iwashita, Tanaka, Mori, Ogasawara, Suga, Koshikawa, Watanabe, Yasuda, Kobayashi, {et~al.}}]{kataoka2024inspire}
Kataoka, J., Iwashita, R., Tanaka, K., {et~al.} 2024, \bibinfo{title}{INSPIRE: Challenge of 50 kg-class satellite to open up MeV gamma-ray astronomy,} Nuclear Instruments and Methods in Physics Research Section A: Accelerators, Spectrometers, Detectors and Associated Equipment, 169518

\bibitem[{J. Kataoka {et~al.}(2025)Kataoka, Ogasawara, Mori, Yamamoto, Joshi, Kojima, Sato, Tanaka, Watanabe, Yasuda, {et~al.}}]{kataoka2025revisiting}
Kataoka, J., Ogasawara, S., Mori, R., {et~al.} 2025, \bibinfo{title}{Revisiting the ARM cut in Compton gamma-ray imaging and its application to the INSPIRE detector,} Journal of Instrumentation, 20, P10009

\bibitem[{W.~S. Paciesas {et~al.}(1999)Paciesas, Meegan, Pendleton, Briggs, Kouveliotou, Koshut, Lestrade, McCollough, Brainerd, Hakkila, {et~al.}}]{paciesas1999fourth}
Paciesas, W.~S., Meegan, C.~A., Pendleton, G.~N., {et~al.} 1999, \bibinfo{title}{The fourth BATSE gamma-ray burst catalog (revised),} The Astrophysical Journal Supplement Series, 122, 465

\bibitem[{T. Piran(2004)Piran}]{piran2004physics}
Piran, T. 2004, \bibinfo{title}{The physics of gamma-ray bursts,} Reviews of modern physics, 76, 1143

\bibitem[{O. Ronneberger {et~al.}(2015)Ronneberger, Fischer, \& Brox}]{ronneberger2015u}
Ronneberger, O., Fischer, P., \& Brox, T. 2015, in Medical image computing and computer-assisted intervention--MICCAI 2015: 18th international conference, Munich, Germany, October 5-9, 2015, proceedings, part III 18, Springer, 234--241

\bibitem[{S. Sato {et~al.}(2024)Sato, Tanaka, \& Kataoka}]{sato2024comptonnet}
Sato, S., Tanaka, K., \& Kataoka, J. 2024, \bibinfo{title}{ComptonNet: A direct reconstruction model for Compton camera,} Applied Physics Letters, 124

\bibitem[{S. Sato {et~al.}(2020)Sato, Kataoka, Kotoku, Taki, Oyama, Tagawa, Fujieda, Nishi, \& Toyoda}]{sato2020high}
Sato, S., Kataoka, J., Kotoku, J., {et~al.} 2020, \bibinfo{title}{High-statistics image generation from sparse radiation images by four types of machine-learning models,} Journal of Instrumentation, 15, P10026

\bibitem[{S. Serjeant {et~al.}(2020)Serjeant, Elvis, \& Tinetti}]{serjeant2020future}
Serjeant, S., Elvis, M., \& Tinetti, G. 2020, \bibinfo{title}{The future of astronomy with small satellites,} Nature Astronomy, 4, 1031

\bibitem[{G.~L. Stephens {et~al.}(2015)Stephens, O'Brien, Webster, Pilewski, Kato, \& Li}]{stephens2015albedo}
Stephens, G.~L., O'Brien, D., Webster, P.~J., {et~al.} 2015, \bibinfo{title}{The albedo of Earth,} Reviews of geophysics, 53, 141

\bibitem[{K. Tanaka {et~al.}(2024)Tanaka, Kataoka, Iwashita, Mori, Suga, Ogasawara, Tao, Yatsu, Toshihiro, Takeda, {et~al.}}]{tanaka2024mev}
Tanaka, K., Kataoka, J., Iwashita, R., {et~al.} 2024, in Space Telescopes and Instrumentation 2024: Ultraviolet to Gamma Ray, Vol. 13093, SPIE, 858--865

\bibitem[{Z. Wang {et~al.}(2004)Wang, Bovik, Sheikh, \& Simoncelli}]{wang2004image}
Wang, Z., Bovik, A.~C., Sheikh, H.~R., \& Simoncelli, E.~P. 2004, \bibinfo{title}{Image quality assessment: from error visibility to structural similarity,} IEEE transactions on image processing, 13, 600

\bibitem[{N. Werner {et~al.}(2018)Werner, {\v{R}}{\'\i}pa, P{\'a}l, Ohno, Tarcai, Torigoe, Tanaka, Uchida, M{\'e}sz{\'a}ros, Galg{\'o}czi, {et~al.}}]{werner2018camelot}
Werner, N., {\v{R}}{\'\i}pa, J., P{\'a}l, A., {et~al.} 2018, in Space Telescopes and Instrumentation 2018: Ultraviolet to Gamma Ray, Vol. 10699, SPIE, 672--686

\bibitem[{S. Woosley \& J. Bloom(2006)Woosley \& Bloom}]{woosley2006supernova}
Woosley, S., \& Bloom, J. 2006, \bibinfo{title}{The supernova--gamma-ray burst connection,} Annu. Rev. Astron. Astrophys., 44, 507

\bibitem[{Z. Yao {et~al.}(2022)Yao, Shi, Tian, Xiao, Geng, \& Tang}]{yao2022rapid}
Yao, Z., Shi, C., Tian, F., {et~al.} 2022, \bibinfo{title}{Rapid and high-resolution deep learning--based radiopharmaceutical imaging with 3D-CZT Compton camera and sparse projection data,} Medical Physics, 49, 7336

\end{thebibliography}
\bibliographystyle{aasjournalv7}



\end{document}